\documentclass[10pt,twocolumn,letterpaper]{article}
\usepackage{}

\usepackage{cvpr}
\usepackage{times}
\usepackage{epsfig}
\usepackage{graphicx}
\usepackage{amsmath}
\usepackage{amssymb}
\usepackage{booktabs}

\usepackage[breaklinks=true,bookmarks=false]{hyperref}
\usepackage{algorithm}
\usepackage{algorithmic}
\usepackage{stfloats}
\usepackage{amsfonts}
\usepackage{slashbox}
\usepackage{pdfpages}

\cvprfinalcopy 

\ifcvprfinal\fi
\begin{document}

\title{Jointly Modeling Embedding and Translation to Bridge Video and Language}

\author{Yingwei Pan $^{\ddag}$, Tao Mei $^{\dag}$, Ting Yao $^{\dag}$, Houqiang Li $^{\ddag}$, and Yong Rui $^{\dag}$ \\
         $^{\dag}$ Microsoft Research, Beijing, China\\
         $^{\ddag}$ University of Science and Technology of China, Hefei, China\\
{\tt\small panyw.ustc@gmail.com, \{tmei, tiyao, yongrui\}@microsoft.com, lihq@ustc.edu.cn}
}

\maketitle
\thispagestyle{empty}

\maketitle
\begin{abstract}
Automatically describing video content with natural language is a fundamental challenge of multimedia. Recurrent Neural Networks (RNN), which models sequence dynamics, has attracted increasing attention on visual interpretation. However, most existing approaches generate a word locally with given previous words and the visual content, while the relationship between sentence semantics and visual content is not holistically exploited. As a result, the generated sentences may be contextually correct but the semantics (e.g., subjects, verbs or objects) are not true.

This paper presents a novel unified framework, named Long Short-Term Memory with visual-semantic Embedding (LSTM-E), which can simultaneously explore the learning of LSTM and visual-semantic embedding. The former aims to locally maximize the probability of generating the next word given previous words and visual content, while the latter is to create a visual-semantic embedding space for enforcing the relationship between the semantics of the entire sentence and visual content. Our proposed LSTM-E consists of three components: a 2-D and/or 3-D deep convolutional neural networks for learning powerful video representation, a deep RNN for generating sentences, and a joint embedding model for exploring the relationships between visual content and sentence semantics. The experiments on YouTube2Text dataset show that our proposed LSTM-E achieves to-date the best reported performance in generating natural sentences: 45.3\% and 31.0\% in terms of BLEU@4 and METEOR, respectively. We also demonstrate that LSTM-E is superior in predicting Subject-Verb-Object (SVO) triplets to several state-of-the-art techniques.
\end{abstract}

\section{Introduction}
Video has become ubiquitous on the Internet, broadcasting channels, as well as personal devices. This has encouraged the development of advanced techniques to analyze the semantic video content for a wide variety of applications. Recognition of videos has been a fundamental challenge of multimedia for decades. Previous research has predominantly focused on recognizing videos with a predefined yet very limited set of individual words. Thanks to the recent development of Recurrent Neural Networks (RNN), researchers have strived to automatically describe video content with a complete and natural sentence, which can be regarded as the ultimate goal of video understanding.

\begin{figure}[!tb]
\centering
{\includegraphics[width=0.45\textwidth]{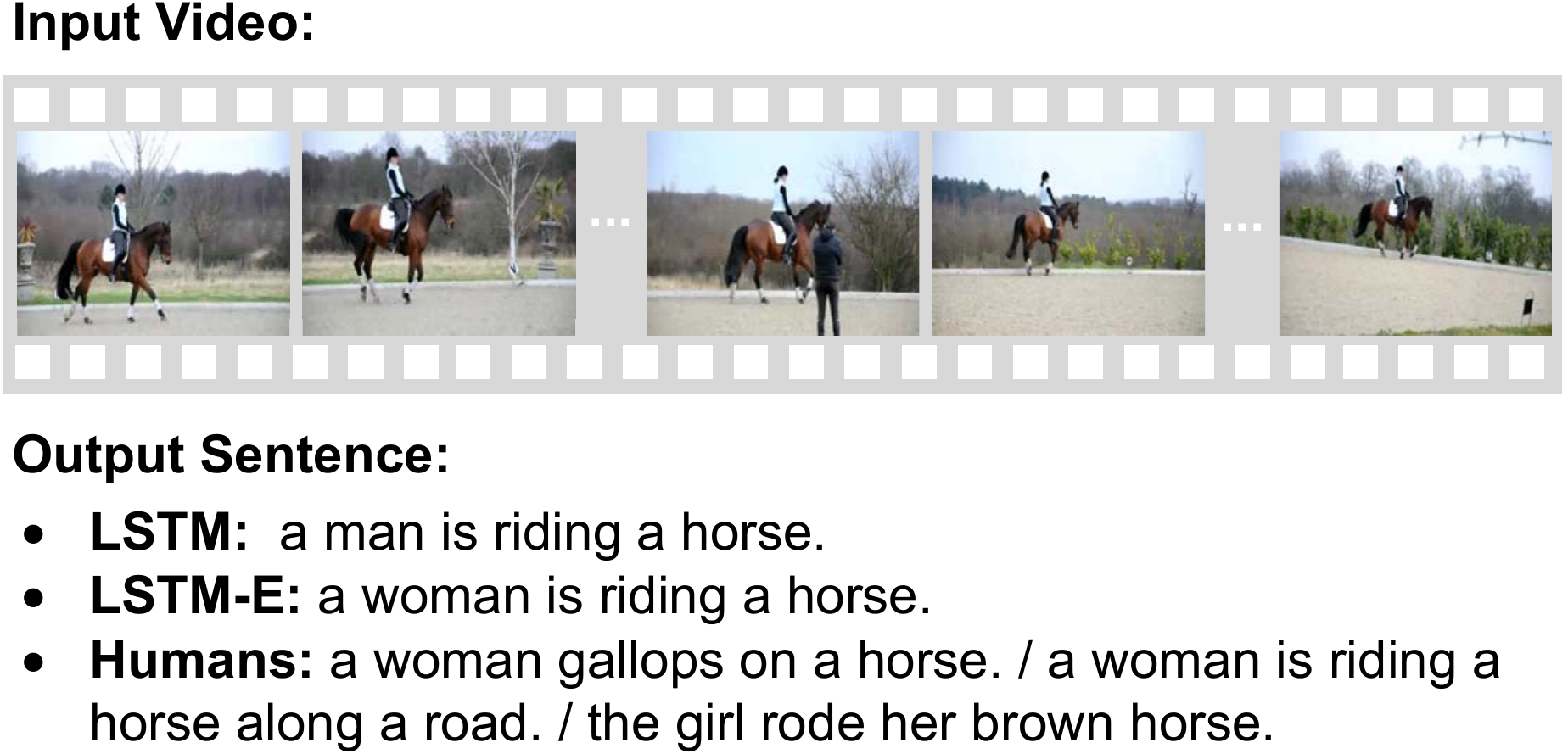}}
\caption{\small Examples of video description generation. Input: a short video. Output: a natural language sentence describing the main content of the input video.}
\label{fig:fig1}
\end{figure}

Figure \ref{fig:fig1} shows the examples of video description generation. Given an input video, the generated sentences are to describe video content, ideally encapsulating its most informative dynamics. There is a wide variety of video applications based on the description, ranging from editing, indexing, search, to sharing. However, the problem itself has been taken as a ground challenge for decades in the research communities, as the description generation model should be powerful enough not only to recognize key objects from visual content, but also discover their spatio-temporal relationships and the dynamics expressed in a natural language as well.

Despite the difficulty of the problem, there have been a few attempts to address video description generation \cite{Donahue14, Venugopalan14, Yao15}, and image caption generation \cite{Fang:CVPR15, Karpathy:CVPR15, Kiros15, Mao:NIPS14, Vinyals14}, which are mainly inspired by recent advances in machine translation using Recurrent Neural Networks (RNN) \cite{Bahdanau14, Sutskever:NIPS14}. The standard RNN is a nonlinear dynamical system that maps sequences to sequences. Although the gradients of the RNN are easy to compute, RNN models are difficult to train, especially when the problems have long-range temporal dependencies, due to the well-known ``vanishing gradient" effect \cite{Bengio:TNN94, Martens:ICML11}. As such, the Long Short-Term Memory (LSTM) model was proposed to overcome the vanishing gradients problem by incorporating memory units, which allow the network to learn when to forget previous hidden states and when to update hidden states \cite{Hochreiter:NC97}. LSTM has been successfully adopted to several tasks, e.g., speech recognition \cite{Graves:ICML14}, language translation \cite{Bahdanau14} and image caption \cite{Mao:NIPS14, Vinyals14}. Thus, we follow this elegant recipe and use LSTM as our RNN model to generate the video sentence in this paper.

Moreover, existing video description generation approaches mainly optimize the next word given the input video and previous words locally, while leaving the relationship between the semantics of the entire sentence and video content unexploited. As a result, the generated sentences can suffer from robustness problem. It is often the case that the output sentence from existing approaches may be contextually correct but the semantics (e.g., subjects, verbs or objects) in the sentence are not true. For example, the sentence generated by LSTM-based model for the video in Figure \ref{fig:fig1} is ``a man is riding a horse," which is correct in logic but the subject ``man" is not relevant to the video content.

\begin{figure*}[!tb]
\centering {\includegraphics[width=0.98\textwidth]{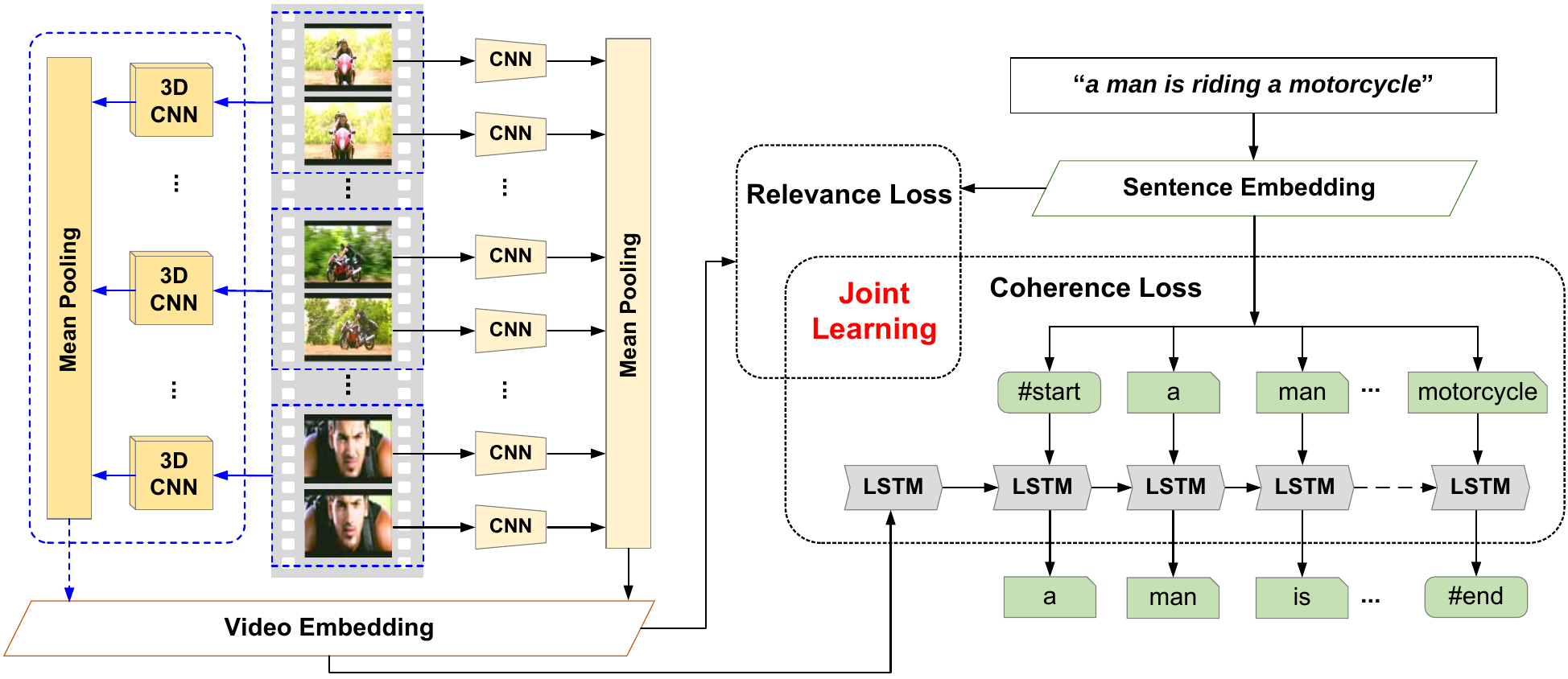}}
\caption{An overview of our LSTM-E framework with a language generating LSTM and a visual-semantic embedding model (better viewed in color). The video representation is produced by mean pooling over the visual features of frames/clips, extracted by a 2-D/3-D CNN. The \emph{relevance loss} is to measure the relationships between the semantics of the entire sentence and video content in the embedding space, while the \emph{coherence loss} is to characterize the contextual relationships among the generated words in the sentence in LSTM. Both LSTM and visual-semantic embedding are jointly learnt by minimizing the two losses.}
\label{fig:fig2}
\end{figure*}

To address the above issues, we leverage the semantics of the entire sentence and visual content to learn a visual-semantic embedding model, which holistically explores the relationships in between. Specifically, we present a novel Long Short-Term Memory with visual-semantic Embedding (LSTM-E) framework to bridge video content and natural language, as shown in Figure \ref{fig:fig2}. Given a video, a 2-D and/or 3-D Convolution Neural Networks (CNN) is utilized to extract visual features of selected video frames/clips, while the video representation is produced by mean pooling over these visual features. Then, a LSTM for generating video sentence and a visual-semantic embedding model are jointly learnt based on the video representation and sentence semantics. The spirit of LSTM-E is to generate video sentence from the viewpoint of mutual reinforcement between \emph{coherence} and \emph{relevance}. \emph{Coherence} expresses the contextual relationships among the generated words with video content which is optimized in LSTM, while \emph{relevance} conveys the relationship between the semantics of the entire sentence and video content which is measured in the visual-semantic embedding.

In summary, this paper makes the following contributions:
\begin{itemize}
  \item We present an end-to-end deep model for automatic video description generation, which incorporates both spatial and temporal structures underlying video.
  \item We propose a novel Long Shot-Term Memory with visual-semantic Embedding (LSTM-E) framework, which considers both the contextual relationship among the words in sentence, and the relationship between the semantics of the entire sentence and video content, for generating natural language of a given video.
  \item The proposed model is evaluated on the popular Youtube2Text corpus and outperforms the-state-of-the-art in terms of both Subject-Verb-Object (SVO) triplet prediction and sentence generation.
\end{itemize}

The remaining parts of the paper are organized as follows. Section \ref{sec:RW} reviews related work. Section \ref{sec:VDG} presents the problem of video description generation, while Section \ref{sec:MET} details our solution of jointly modeling embedding and translation. In Section \ref{sec:EX}, we provide empirical evaluations, followed by the discussions and conclusions in Section \ref{sec:CON}.

\section{RELATED WORK}\label{sec:RW}
There are mainly two directions for translation from visual content. The first direction predefines the special rule for language grammar and split sentence into several parts (e.g. subject, verb, object). With such sentence fragments, many works align each part with visual content and then generate the sentence for corresponding visual content: \cite{Kulkarni:PAMI13} use Conditional Random Field (CRF) model to produce sentence for image and in \cite{Farhadi:ECCV10}, a Markov Random Field (MRF) model is proposed to attach a descriptive sentence to the given image. For video translation, Rohrbach \emph{et al.} \cite{Rohrbach:ICCV13} learn a CRF to model the relationships between different components of the input video and generate descriptions for video. Guadarrama \emph{et al.} \cite{Guadarrama:ICCV13} use semantic hierarchies to choose an appropriate level of the specificity and accuracy of sentence fragments. This direction is highly depended on the templates of sentence and can only generate sentence with syntactical structure.

Another direction is to learn the probability distribution in the common space of visual content and textual sentence. In this direction, several works explore such probability distribution using topic models \cite{Barnard:JMLR03, Jia:ICCV11} and neural networks \cite{Donahue14, Kiros:ICML14, Mao:NIPS14, Venugopalan15, Venugopalan14, Vinyals14, Yao15}. They can generate sentence more flexibly. Most recently, several methods have been proposed for visual to sentence task based on the neural networks and most of them are utilizing the RNN due to its successful use in sequence to sequence learning for machine translation \cite{Bahdanau14, Sutskever:NIPS14}. Kiros \emph{et al.} \cite{Kiros:ICML14} firstly take the neural networks to generate sentence for image by proposing a image-text multimodal log-bilinear neural language model. In another work by Mao \emph{et al.} \cite{Mao:NIPS14}, a multimodal Recurrent Neural Networks (m-RNN) model is proposed for image to caption, which directly models the probability of generating a word given previous words and image. In \cite{Vinyals14}, Vinyals \emph{et al.} propose an end-to-end neural networks system by utilizing LSTM to generate sentence for image. For video translation, an end-to-end LSTM based model is proposed in \cite{Venugopalan14}, which only reads the sequence of video frames and then generates a natural sentence. The model is further extended by inputting both frames and optical flow in \cite{Venugopalan15}. Yao \emph{et al.} propose to use a 3-D convolutional neural networks for modeling video clip dynamic temporal structure and an attention mechanism to select the most relevant temporal clips \cite{Yao15}. Then, the resulting video representations are fed into the text-generating RNN.

Our work belongs to the second direction. However, most of the above approaches in this direction mainly focus on optimizing the contextual relationship among words to generate sentence given visual content, while the relationship between the semantics of the entire sentence and visual content is not fully explored. Our work is different that we claim to generate video sentence by jointly exploiting the two relationships, which characterize the complementary properties of \emph{coherence} and \emph{relevance} of a generated sentence, respectively.

\section{Video Description Generation}\label{sec:VDG}
Our goal is to generate language sentences for videos. What makes a good sentence? Beyond describing important persons, objects, scenes, and actions by words, it must also convey how one word leads to the next. Specifically, we define a good sentence as a coherent chain of words in which each word influences the next through contextual information. Furthermore, the semantics of the entire sentence must be relevant to the video content. We begin this Section by presenting the problem formulation, and followed by the proposal of two losses on measuring \emph{coherence} and \emph{relevance}, respectively.

\subsection{Problem Formulation}
Suppose we have a video $\mathcal {V}$ with $N_v$ sample frames/clips (uniform sampling) to be described by a textual sentence $\mathcal {S}$, where $\mathcal{S} = \{w_1, w_2, ..., w_{N_s}\}$ consisting of $N_s$ words. Let ${\bf{v}}\in {\mathbb{R}}^{D_v}$ and ${\bf{w}}_t\in {{\mathbb{R}}^{D_w}}$ denote the $D_v$-dimensional visual features of a video $\mathcal {V}$ and the $D_w$-dimensional textual features of the $t$-th word in sentence $\mathcal{S}$, respectively. As a sentence consists of a sequence of words, a sentence can be represented by a $D_w\times N_s$ matrix $\bf{W}\equiv [{\bf{w}}_1, {\bf{w}}_2,...,{\bf{w}}_{N_s}]$, with each word in the sentence as its column vector. Furthermore, we denote another feature vector $\bf{s}$ in the text space for representing a sentence as a whole.

In the video description generation problem, on one hand, the generated descriptive sentence must be able to depict the main contents of a video precisely, and on the other, the words in the sentence should be organized coherently in language. Therefore, we can formulate the video description generation problem by minimizing the following energy loss function
\begin{equation}\label{Eq:Eq1}
E(\mathcal{V}, \mathcal{S}) = (1 - \lambda) \times E_r({\bf{v}, \bf{s}}) + \lambda \times E_c(\bf{v}, \bf{W})~,
\end{equation}
where $E_r(\bf{v}, \bf{s})$ and $E_c(\bf{v}, \bf{W})$ are the \emph{relevance loss} and \emph{coherence loss}, respectively. The former measures the relevance degree of the video content and sentence semantics and we build an visual-semantic embedding for this purpose, which is introduced in Section \ref{ssec:VSE}. The latter estimates the contextual relationships among the generated words in the sentence and we use LSTM-based RNN as our model, which is presented in Section \ref{ssec:TSL}. The tradeoff between these two competing losses is captured by linear fusion with a positive parameter $\lambda$.

\subsection{Visual-Semantic Embedding}\label{ssec:VSE}
In order to effectively represent the visual content of a video, we first use a 2-D and/or 3-D deep convolutional neural networks (CNN), which is powerful to produce a rich representation of each sampled frame/clip from the video. Then, we perform ``mean pooling" process over all the frames/clips to generate a single $D_v$-dimension vector $\bf{v}$ for each video $\mathcal {V}$. The sentence feature $\bf{s}$ is produced by the feature vectors ${\bf{w}}_t (t=1,2,...,N_s)$ of each word in the sentence. We first encode each word $w_t$ as ``one-hot" vector (binary index vector in a vocabulary), thus the dimension of feature vector ${\bf{w}}_t$, i.e. $D_w$, is the vocabulary size. Then the binary TF weights are calculated over all words of the sentence to produce the integrated representation of the entire sentence, denoted by ${\bf{s}} \in {\mathbb{R}}^{D_w}$, with the same dimension of~${\bf{w}}_t$.

We assume that a low-dimensional embedding exists for the representation of video and sentence. The linear mapping function can be derived from this embedding by
\begin{equation}\label{Eq:Eq2}
{{\bf{v}}_e} = {{\bf{T}}_v}{\bf{v}}  ~~{\rm{and}}~~  {{\bf{s}}_e} = {{\bf{T}}_s}{\bf{s}},
\end{equation}
where $D_e$ is the dimensionality of the embedding, and ${{\bf{T}}_v} \in {{\mathbb{R}}^{{D_e} \times {D_v}}}$ and ${{\bf{T}}_s} \in {{\mathbb{R}}^{{D_e} \times {D_s}}}$ are the transformation matrices that project the video content and semantic sentence into the common embedding, respectively.

To measure the relevance between the video content and semantic sentence, one natural way is to compute the distance between their mappings in the embedding. Thus, we define the relevance loss as
\begin{equation}\label{Eq:Eq4}
E_r({\bf{v}}, {\bf{s}}) = \left\| {{{\bf{T}}_v}{\bf{v}} - {{\bf{T}}_s}{\bf{s}}} \right\|_2^2.
\end{equation}

We strengthen the relevance between video content and semantic sentence by minimizing the relevance loss. As such, the generated sentence is expected to better manifest the semantics of videos.

\subsection{Translation by Sequence Learning}\label{ssec:TSL}
Inspired by the recent successes of probabilistic sequence models leveraged in statistical machine translation \cite{Donahue14, Vinyals14}, we define our coherence loss as
\begin{equation}\label{Eq:Eq5}
E_c({\bf{v}}, {\bf{W}}) = -\log {\Pr{({\bf{W}}|{\bf{v}})}}.
\end{equation}

Assuming that a generative model of $\bf{W}$ that produces each word in the sequence in order, the $\log$ probability of the sentence is given by the sum of the $\log$ probabilities over the word and can be expressed as:
\begin{equation}\label{Eq:Eq6}
\log {\Pr{({\bf{W}}|{\bf{v}})}} =  \sum\limits_{t = 0}^{{N_s}} {\log \Pr\left( {\left. {{{\bf{w}}_t}} \right|{{\bf{v}}},{{\bf{w}}_0}, \ldots ,{{\bf{w}}_{t - 1}}} \right)}.
\end{equation}
By minimizing the coherence loss, the contextual relationship among the words in the sentence can be guaranteed, making the sentence coherent and smooth.

In video description generation task, both the relevance loss and coherence loss need to be estimated to complete the whole energy function. We will present a solution to jointly model the two losses in a deep recurrent neural networks in the next sections.

\section{Joint Modeling Embedding and Translation}\label{sec:MET}
Following the \emph{relevance} and \emph{coherence} criteria, this work proposes a Long Short-Term Memory with visual-semantic Embedding (LSTM-E) model for video description generation. The basic idea of LSTM-E is to translate the video representation from a 2-D and/or 3-D deep convolutional network to the desired output sentence by using LSTM-type RNN model. Figure \ref{fig:fig2} shows an overview of LSTM-E model. In particular, the training of LSTM-E is performed by simultaneously minimizing the \emph{relevance loss} and \emph{coherence loss}. Therefore, the formulation presented in Eq.(\ref{Eq:Eq1}) is equivalent to minimizing the following energy function
\begin{equation}\label{Eq:Eq7}
\begin{array}{l}
E(\mathcal{V}, \mathcal{S}) = (1-\lambda) \times \left\| {{{\bf{T}}_v}{\bf{v}} - {{\bf{T}}_s}{\bf{s}}} \right\|_2^2 - \\
~~~~~~~~~~~\lambda \times \sum\limits_{t = 0}^{{N_s}} {\log \Pr\left( {\left. {{{\bf{w}}_t}} \right|{{\bf{v}}},{{\bf{w}}_0}, \ldots ,{{\bf{w}}_{t - 1}}}; \theta; {{\bf{T}}_v}; {{\bf{T}}_s} \right)}
\end{array},
\end{equation}
where $\theta$ are the parameters of our LSTM-E models.

In the following, we will first present the architecture of LSTM memory cell, followed by jointly modeling with visual-semantic embedding.

\subsection{Long Short Term Memory}\label{ssec:LSTM}
We briefly introduce the standard Long Short-Term Memory (LSTM) \cite{Hochreiter:NC97}, a variant of RNN, which can capture long-term temporal information by mapping input sequences to a sequence of hidden states and then hidden states to outputs. To address the vanishing gradients problem in traditional RNN training, LSTM incorporates a memory cell which can maintain its states over time and non-linear gating units which control the information flow into and out of the cell. As much light has been threw on LSTM recently, many improvements have been made to the LSTM architecture on its original formulation \cite{Hochreiter:NC97}. We adopt the LSTM architecture as described in \cite{Zaremba14}, which omits the peephole connections in previous work \cite{Graves:NN05}.

\begin{figure}[!tb]
\centering {\includegraphics[width=0.38\textwidth]{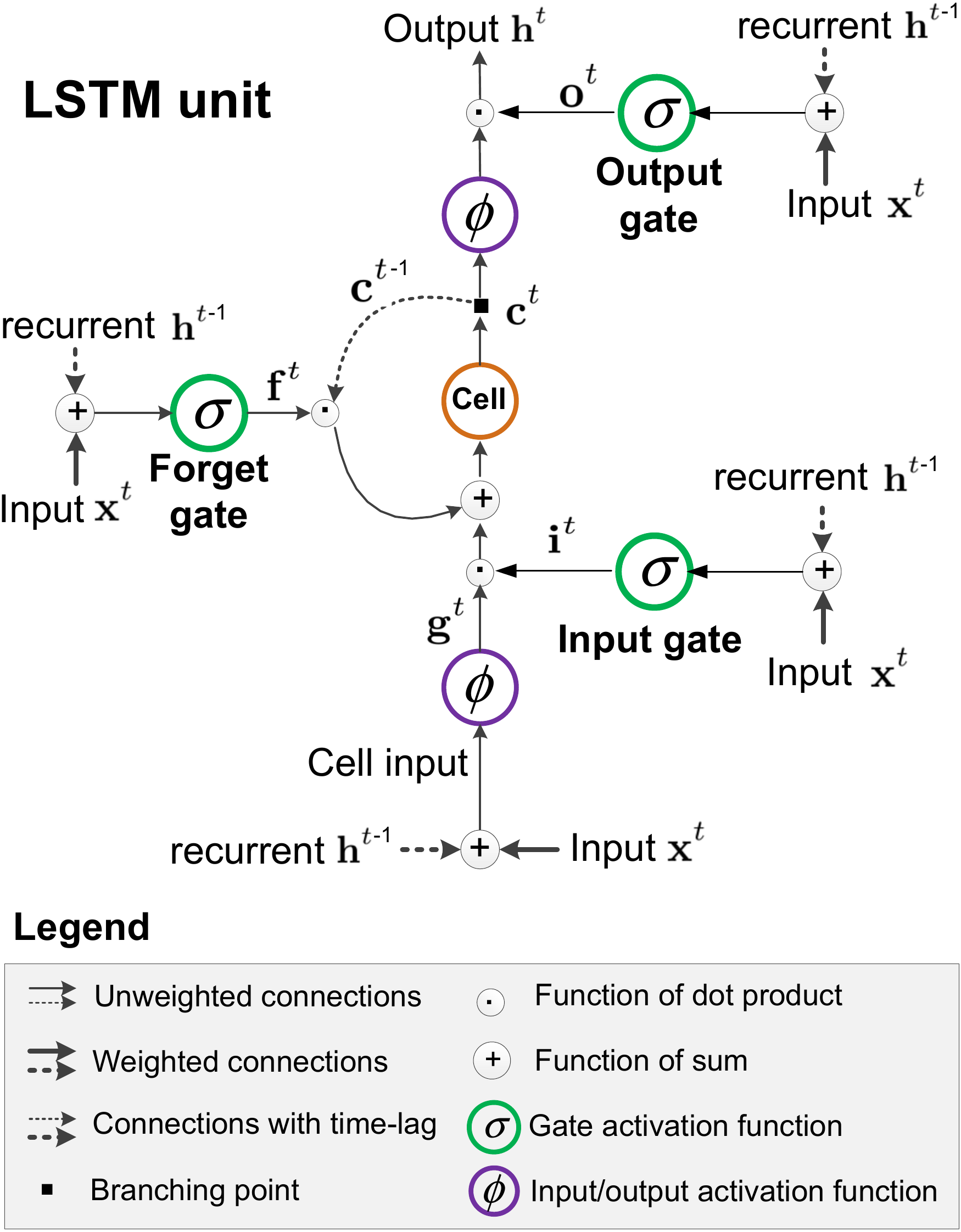}}
\caption{A diagram of an LSTM memory cell.}
\label{fig:fig3}
\end{figure}

A diagram of the LSTM unit can be seen in Figure \ref{fig:fig3}. It consists of a single memory cell, an input activation function, an output activation function, and three gates (input, forget and output). The hidden state of the cell is recurrently connected back to the input and three gates. The memory cell updates its hidden state by combining the previous cell state which is modulated by the forget gate and a function of the current input and the previous output, modulated by the input gate. The forget gate is a critical component of the LSTM unit, which can control what to be remembered and what to be forgotten by the cell and somehow can avoid the gradient from vanishing or exploding when back propagating through time. Having been updated, the cell state is mapped to (-1,1) range through an output activation function which is necessary whenever the cell state is unbounded. Finally, the output gate determines how much of the memory cell flows into the output. These additions to the single memory cell enable LSTM to capture extremely complex and long-term temporal dynamics which is impossible for traditional RNN.

The vector formulas for a LSTM layer forward pass are given below. For timestep $t$, ${\bf{x}}^t$ and ${\bf{h}}^t$ are the input and output vector respectively, $\bf{T}$ are input weights matrices, $\bf{R}$ are recurrent weight matrices and $\bf{b}$ are bias vectors. $Logic\ sigmoid$ $\sigma (x) = \frac{1}{{1 + {e^{ - x}}}}$ and $hyperbolic\ tangent$ $\phi (x) = \frac{{{e^x} - {e^{ - x}}}}{{{e^x} + {e^{ - x}}}}$ are element-wise non-linear activation functions, mapping real numbers to $(0,1)$ and $(-1,1)$ separately. The dot product and sum of two vectors are denoted with $\odot$ and $\textcircled{+}$, respectively. Given inputs ${\bf{x}}^t$, ${\bf{h}}^{t-1}$ and ${\bf{c}}^{t-1}$, the LSTM unit updates for timestep $t$ are:
\begin{align*}
&{{\bf{g}}^t} = \phi ({{\bf{T}}_g}{{\bf{x}}^t} + {{\bf{R}}_g}{{\bf{h}}^{t - 1}} + {{\bf{b}}_g}) &cell\ input\\
&{{\bf{i}}^t} = \sigma ({{\bf{T}}_i}{{\bf{x}}^t} + {{\bf{R}}_i}{{\bf{h}}^{t - 1}} + {{\bf{b}}_i}) &input\ gate \\
&{{\bf{f}}^t} = \sigma ({{\bf{T}}_f}{{\bf{x}}^t} + {{\bf{R}}_f}{{\bf{h}}^{t - 1}} + {{\bf{b}}_f}) &forget gate \\
&{{\bf{c}}^t} = {\bf{g}}^t \odot {\bf{i}}^t + \bf{c}^{t-1} \odot {\bf{f}}^t &cell\ state \\
&{{\bf{o}}^t} = \sigma ({{\bf{T}}_o}{{\bf{x}}^t} + {{\bf{R}}_o}{{\bf{h}}^{t - 1}} + {{\bf{b}}_o}) &output\ gate \\
&{{\bf{h}}^t} = \phi ({\bf{c}}^t) \odot {\bf{o}}^t &cell\ output \\
\end{align*}

\subsection{LSTM with Visual-Semantic Embedding}\label{ssec:RNNE}
By further incorporating a visual-semantic embedding, our LSTM-E architecture is to jointly model embedding and translation. In the training stage, given the video-sentence pair, the inputs of LSTM are the representations of the video and the words in the sentence after mapping into the embedding. As mentioned above, here we train the LSTM model to predict each word in the sentence given the embedding of visual feature for video and previous words. There are multiple ways that can be used to combine the visual content and words in LSTM unit updating procedure. The first one is to feed the visual content at each time step as an extra input for LSTM to emphasize the visual content frequently among LSTM memory cells. The second one only inputs the visual content once at the initial step to inform the whole memory cells in LSTM about the visual content. As empirically verified in \cite{Vinyals14}, feeding the image at each time yields inferior results, due to the fact that the network can explicitly exploit noise and overfits more easily. Therefore, we adopt the second approach to arrange the inputs into LSTM in our architecture. Given the video ${\bf{v}}$ and its corresponding sentence ${\bf{W}}\equiv [{\bf{w}}_0, {\bf{w}}_1,...,{\bf{w}}_{N_s}]$, the LSTM updating procedure is as following:
\begin{equation}\label{Eq:Eq8}
{{\bf{x}}^{ - 1}} = {{\bf{T}}_v}{\bf{v}}\qquad\qquad\qquad\quad
\end{equation}
\begin{equation}\label{Eq:Eq9}
{{\bf{x}}^t} = {{\bf{T}}_s}{{\bf{w}}_t},t \in \left\{ {0, \ldots ,{N_s}-1} \right\}
\end{equation}
\begin{equation}\label{Eq:Eq10}
{{\bf{h}}^{t}} = f\left( {{{\bf{x}}^t}} \right),t \in \left\{ {0, \ldots ,{N_s}-1} \right\}
\end{equation}
where $f$ is the updating function within LSTM unit. Please note that for the input sentence ${\bf{W}} \equiv \left\{ {{{\bf{w}}_0}, \ldots,{{\bf{w}}_{{N_s}}}} \right\}$, we take ${{\bf{w}}_0}$ as the start sign word to inform the beginning of sentence and ${{\bf{w}}_{{N_s}}}$ as the end sign word which indicates the end of sentence, both of the special sign words are included in our vocabulary. Most specifically, at the initial time step, the video representation in the embedding is set as the input for LSTM, and then in the next steps, word embedding ${{\bf{x}}^t}$ will be input into the LSTM along with the previous step's hidden state ${{\bf{h}}^{t-1}}$. In each time step (except the initial step), we use the LSTM cell output ${{\bf{h}}^{t}}$ to predict the next word. Here a softmax layer is applied after the LSTM layer to produce a probability distribution over all the ${D_s}$ words in the vocabulary as
\begin{equation}\label{Eq:Eq11}
{{\Pr}_{t+1}}\left( {{w_{t+1}}} \right) = \frac{{\exp \left( {{{\bf{T}}_h^{\left( w_{t+1} \right)}}{{\bf{h}}^t}} \right)}}{{\sum\limits_{w \in {\mathcal {W}}} {\exp \left( {{{{\bf{T}}_h^{\left( w \right)}}}{{\bf{h}}^t} } \right)} }},
\end{equation}
where $\mathcal {W}$ is the word vocabulary space, ${{\bf{T}}_h^{(w)}}$ is the parameter matrix in softmax layer. Therefore, we can obtain the next word based on such probability distribution until the end sign word is emitted.

Accordingly, we define our loss function as follows:
\begin{equation}\label{Eq:Eq12}
\begin{array}{l}
E(\mathcal{V}, \mathcal{S}) = (1-\lambda) \times \left\| {{{\bf{T}}_v}{\bf{v}} - {{\bf{T}}_s}{\bf{s}}} \right\|_2^2 - \\
~~~~~~~~~~~\lambda \times \sum\limits_{t = 1}^{{N_s}} {\log \Pr_t({\bf{w}}_t )}
\end{array}.
\end{equation}

Let $N$ denote the number of video-sentence pairs in the training dataset, we have the following optimization problem:
\begin{equation}\label{Eq:Eq13}
\begin{array}{l}
\mathop {\min }\limits_{{{\bf{T}}_v},{{\bf{T}}_s},{{\bf{T}}_h},\theta } \frac{1}{N}\sum\limits_{i = 1}^N {\left( E(\mathcal{V}^{(i)}, \mathcal{S}^{(i)})  \right)} \\
~~~~~~~~~~~~~+ \left\| {{{\bf{T}}_v}} \right\|_2^2 + \left\| {{{\bf{T}}_s}} \right\|_2^2 + \left\| {{{\bf{T}}_h}} \right\|_2^2 + \left\| \theta  \right\|_2^2
\end{array},
\end{equation}
where the first term is the combination of the \emph{relevance loss} and \emph{coherence loss}, while the rest are regularization terms for video embedding, sentence embedding, softmax layer and LSTM, respectively.

The above overall objective is optimized over the whole training video-sentence pairs using stochastic gradient descent. By minimizing this objective function, our LSTM-E model takes into account both the contextual relationships among the words in sentence (\emph{coherence}) and the relationships between the semantics of the entire sentence and video content (\emph{relevance}).

For sentence generation, there are two common strategies to translate the given video. The first approach is to sample the next word from the probability distribution at each timestep and set its representation in embedding space as the LSTM input for next timestep until the end sign word is sampled or the maximum sentence size is reached. Another method is select the top-$k$ best sentence for each timestep and sets them as the candidates for next timestep based on which to generate new top-$k$ best sentence. To make the generation process concise and efficient, we adopt the similar way as the latter one but set $k$ as 1. Therefore, at each timestep, we choose the word with maximum probability as the predicted word and input its embedded feature in the next timestep until the model outputs the end sign word.

\section{Experiments}\label{sec:EX}
In this section, we will first introduce our experimental setting. Then, the evaluation results compared with state-of-the-arts on two tasks, i.e., Subject-Verb-Object (SVO) triplet prediction and natural sentence generation tasks, are reported. Finally, the effect of tradeoff parameter between \emph{coherence} and \emph{relevance} and the size of hidden layer in LSTM are presented.

\subsection{Experimental Setting}\label{ssec:Set}
We conduct our experiments mainly on the Microsoft Research Video Description Corpus (YouTube2Text) \cite{Chen:ACL11}, which have been used in several prior works \cite{Guadarrama:ICCV13, Thomason:COLING14, Xu:AAAI15} on action recognition and video description generation tasks. This video corpus contains 1,970 YouTube snippets which cover a wide range of daily activities such as ``people doing exercises," ``playing music," and ``cooking." We use the roughly 40 available English descriptions per video. In our experiments, following the setting used in prior works on video description generation \cite{Guadarrama:ICCV13, Xu:AAAI15}, we pick 1,200 videos to be used as training data, 100 videos for validation and 670 videos for testing.

We compare our LSTM-E architecture with two 2-D CNN of AlexNet \cite{Alex:NIPS12} and the 19-layer VGG \cite{Simonyan14} network both pre-trained on Imagenet ILSVRC12 dataset \cite{ILSVRC15}, and one 3-D CNN of C3D \cite{Tran15} pre-trained on Sports-1M video dataset \cite{KarpathyCVPR14}. Specifically, we take the output of 4096-way fc7 layer from AlexNet, 4096-way fc6 layer from the 19-layer VGG, and 4096-way fc6 layer from C3D as the frame/clip representation, respectively. The dimensionality of the visual-semantic embedding space and the size of hidden layer in LSTM are both set to 512. The tradeoff parameter $\lambda$ leveraging the relevance loss and coherence loss is empirically set to 0.7. The sensitivity of $\lambda$ will be discussed in Section \ref{sssec:TPS}.

\subsection{Performance Comparison}\label{ssec:perf}
We empirically verify the merit of our LSTM-E model from two aspects: SVO triplet prediction and sentence generation for the video-language translation.

\subsubsection{Compared Approaches}\label{sssec:CApp}
To fully evaluate our model, we compare our LSTM-E models with the following non-trivial baseline methods.
\begin{itemize}
  \item Conditional Random Field (CRF) \cite{Xu:AAAI15}: CRF model is developed to incorporate subject-verb and verb-object pairwise relationship based on the word pairwise co-occurrence statistics in the sentence pool.
  \item Canonical Correlation Analysis (CCA) \cite{Socher:CVPR10}: CCA is to build the video-language joint space and generate the SVO triplet by k-nearest-neighbors search in the sentence pool.
  \item Factor Graph Model (FGM) \cite{Thomason:COLING14}: FGM combines knowledge mined from text corpora with visual confidence using a factor graph and performs probabilistic inference to determine the most likely SVO triplets.
  \item Joint Embedding Model (JEM) \cite{Xu:AAAI15}: Proposed most recently, JEM jointly models video and the corresponding text sentences by minimizing the distance of the deep video and compositional text in the joint space.
  \item Long Shot-Term Memory (LSTM): LSTM attempts to directly translate from video pixels to natural language with a single deep neural network. The video representation is by performing mean pooling over the features of frames using AlexNet.
  \item Soft-Attention (SA) \cite{Yao15}: SA combines the frame representation from GoogleNet \cite{Szegedy14} and video clip representation based on a 3-D ConvNet trained on Histograms of Oriented Gradients (HOG), Histograms of Optical Flow (HOF), and Motion Boundary Histogram (MBH) hand-crafted descriptors. Furthermore, a weighted attention mechanism is used to dynamically attend to specific temporal regions of the video while generating sentence.
  \item Sequence to Sequence - Video to Text (S2VT) \cite{Venugopalan15}: S2VT incorporates both RGB and optical flow inputs, and the encoding and decoding of the inputs and word representations are learnt jointly in a parallel manner.
  \item Long Shot-Term Memory with visual-semantic Embedding (LSTM-E): We design four runs for our proposed approach, i.e., LSTM-E (Alex), LSTM-E (VGG), LSTM-E (C3D), and LSTM-E (VGG+C3D). The input frame/clip features of the first three runs are from AlexNet, VGG and C3D network respectively. The input of the last one is to concatenate the features from VGG and C3D.
\end{itemize}

\begin{table}
\centering
\caption{SVO accuracy: Binary accuracy of SVO triplet prediction. We extract SVO triplets from sentences output by LSTM and LSTM-E using a dependency parser.}
\label{table:svo}
\begin{tabular}{l|c|c|c}\hline
~~\textbf{Model}&~~~~\textbf{S\%}~~~~&~~~~\textbf{V\%}~~~~&~~~~\textbf{O\%}~~~~\\ \hline
~~\textbf{FGM}	 &	76.42 &	21.34 &	12.39 \\
~~\textbf{CRF}	 &	77.16 &	22.54 &	9.25 \\
~~\textbf{CCA}	 &	77.16 & 21.04 &	10.99 \\
~~\textbf{JEM}	 &	78.25 & 24.45 & 11.95 \\ \hline
~~\textbf{LSTM} &	71.19 &	19.40 &	9.70 \\ \hline
~~\textbf{LSTM-E (Alex)} & 78.66 & 24.78 & 10.30  \\
~~\textbf{LSTM-E (VGG)} & 80.30  & 27.91 & 12.54  \\
~~\textbf{LSTM-E (C3D)} & 77.31  & 28.81 & 12.39  \\
~~\textbf{LSTM-E (VGG+C3D)} & \textbf{80.45} & \textbf{29.85} & \textbf{13.88}  \\ \hline
\end{tabular}
\end{table}

\subsubsection{SVO Triplet Prediction Task}\label{sssec:SVO}
As SVO triples can capture the compositional semantics of videos, predicting SVO triplet could indicate the quality of a translation system to a large extent.

We adopt SVO accuracy \cite{Xu:AAAI15} which measures the exactness of SVO words by binary (0-1 loss), as the evaluation metric. Table \ref{table:svo} details SVO accuracy of compared seven models. Within these models, the former four models (called \emph{Item driven models}) explicitly optimize to identify the best subject, verb and object items for a video; while the later five models (named \emph{Sentence driven models}) focus on training on objects and actions jointly in a sentence and learn to interpret these in different contexts. For the later five sentence driven models, we extract the SVO triplets from the generated sentences by Stanford Parser\footnote{http://nlp.stanford.edu/software/lex-parser.shtml} and the words are also stemmed. Overall, the results across SVO triplet indicate that almost all the four \emph{Item driven models} exhibit better performance than LSTM model which predicts the next word by only considering the contextual relationships with the previous words given the video content. By jointly modeling the relevance between the semantics of the entire sentence and video content with LSTM, LSTM-E significantly improves LSTM. Furthermore, the performances of LSTM-E (VGG), LSTM-E (C3D), and LSTM-E (VGG+C3D) on Subject, Verb and Object are all above that of the four \emph{Item driven models}. The result basically indicates the advantage of further exploring the relevance holistically between the semantics of the entire sentence and video content in addition to LSTM.

Compared to LSTM-E (Alex), LSTM-E (VGG) using a more powerful frame representation brought by a deeper CNN exhibits significantly better performance. In addition, LSTM-E (C3D) which has a better ability in encapsulating temporal information leads to better performance than LSTM-E (VGG) in terms of Verb prediction accuracy. When combining the features from VGG and C3D, LSTM-E (VGG+C3D) further increases the performance gains.

\begin{table*}
\centering
\caption{BLEU@$N$ and METEOR scores for comparing the quality of the sentence generation. All values are reported as percentage (\%).}
\label{table:bleu}
\begin{tabular}{l|c|c|c|c|c}\hline
~~\textbf{Model}&~~~~\textbf{METEOR}~~~~&~~~~\textbf{BLEU@1}~~~~&~~~~\textbf{BLEU@2}~~~~&~~~~\textbf{BLEU@3}~~~~&~~~~\textbf{BLEU@4}~~~~\\ \hline
~~\textbf{LSTM} &26.9 & 69.8  &	53.3   & 42.1  & 31.2 \\
~~\textbf{SA} &29.6   & \textbf{80.0}   &	64.7   & 52.6  & 42.2  \\
~~\textbf{S2VT} &29.8  &-   &-    &-      &- \\\hline
~~\textbf{LSTM-E (Alex)} &28.3  & 74.5	&59.8 	&49.3 	&38.9	  \\
~~\textbf{LSTM-E (VGG)} &29.5  & 74.9	&60.9 	&50.6 	&40.2	  \\
~~\textbf{LSTM-E (C3D)} &29.9  & 75.7	&62.3 	&52 	&41.7	  \\
~~\textbf{LSTM-E (VGG+C3D)} &\textbf{31.0} &78.8 &\textbf{66.0}  &\textbf{55.4} & \textbf{45.3}\\\hline
\end{tabular}
\end{table*}

\subsubsection{Sentence Generation Task}\label{sssec:SGT}
For \emph{item driven models} including FGM, CRF, CCA and JEM, the sentence generation is often performed by leveraging a series of simple sentence templates (or special language trees) on the SVO triplets \cite{Venugopalan14}. Having verified in \cite{Venugopalan14}, using LSTM architecture can lead to a large performance boost against the template-based sentence generation. Thus, Table \ref{table:bleu} only shows comparisons of LSTM-based sentence generations. We use the BLEU@$N$ \cite{Papineni:ACL02} and METEOR scores \cite{Banerjee:ACL05} against all ground truth sentences. Both metrics have been shown to correlate well with human judgement, and widely used in machine translation literature. Specifically, BLEU@$N$ measures the fraction of $N$-gram (up to 4-gram) that are in common between a hypothesis and a reference or set of references, while METEOR computes unigram precision and recall, extending exact word matches to include similar words based on WordNet synonyms and stemmed tokens. As shown in the Table \ref{table:bleu}, the qualitative results across different $N$ of BLEU and METEOR consistently indicate that the LSTM-E (Alex) significantly outperforms the traditional LSTM model. Moreover, we can find that the performance gain of BLEU@$N$ becomes larger when $N$ increases, where $N$ measures the length of the contiguous sequence in the sentence. This again confirms that LSTM-E is benefited from the way of holistically exploring the relationships between the semantics of the entire sentence and video content by minimizing the distance of their mappings in a visual-semantic embedding. Similar to the observations in SVO prediction task, our LSTM-E (VGG) outperforms LSTM-E (Alex) and can reach 29.5\% METEOR. Furthermore, LSTM-E (C3D) achieves 29.9\% METEOR and improves the performance to 31.0\% when combined with VGG, which makes the improvement over the current two state-of-the-art methods SA by 4.7\% and S2VT by 4.0\%, respectively.

\begin{figure}[!tb]
\centering {\includegraphics[width=0.35\textwidth]{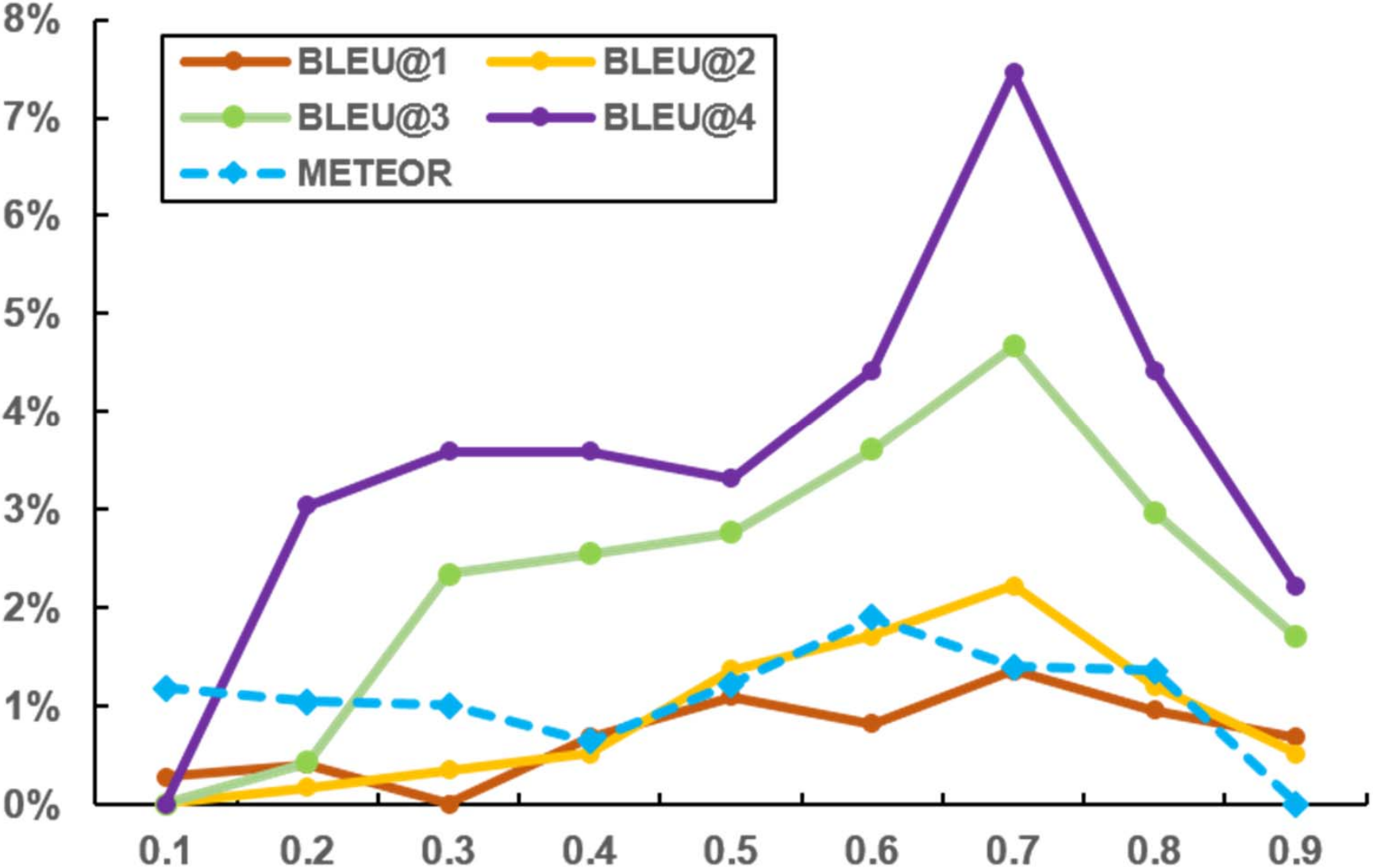}}
\caption{The effect of the tradeoff parameter $\lambda$ measured by BLEU@$N$ and METEOR.}
\label{fig:fig5}
\end{figure}

\begin{figure*}[!tb]
\centering {\includegraphics[width=0.96\textwidth]{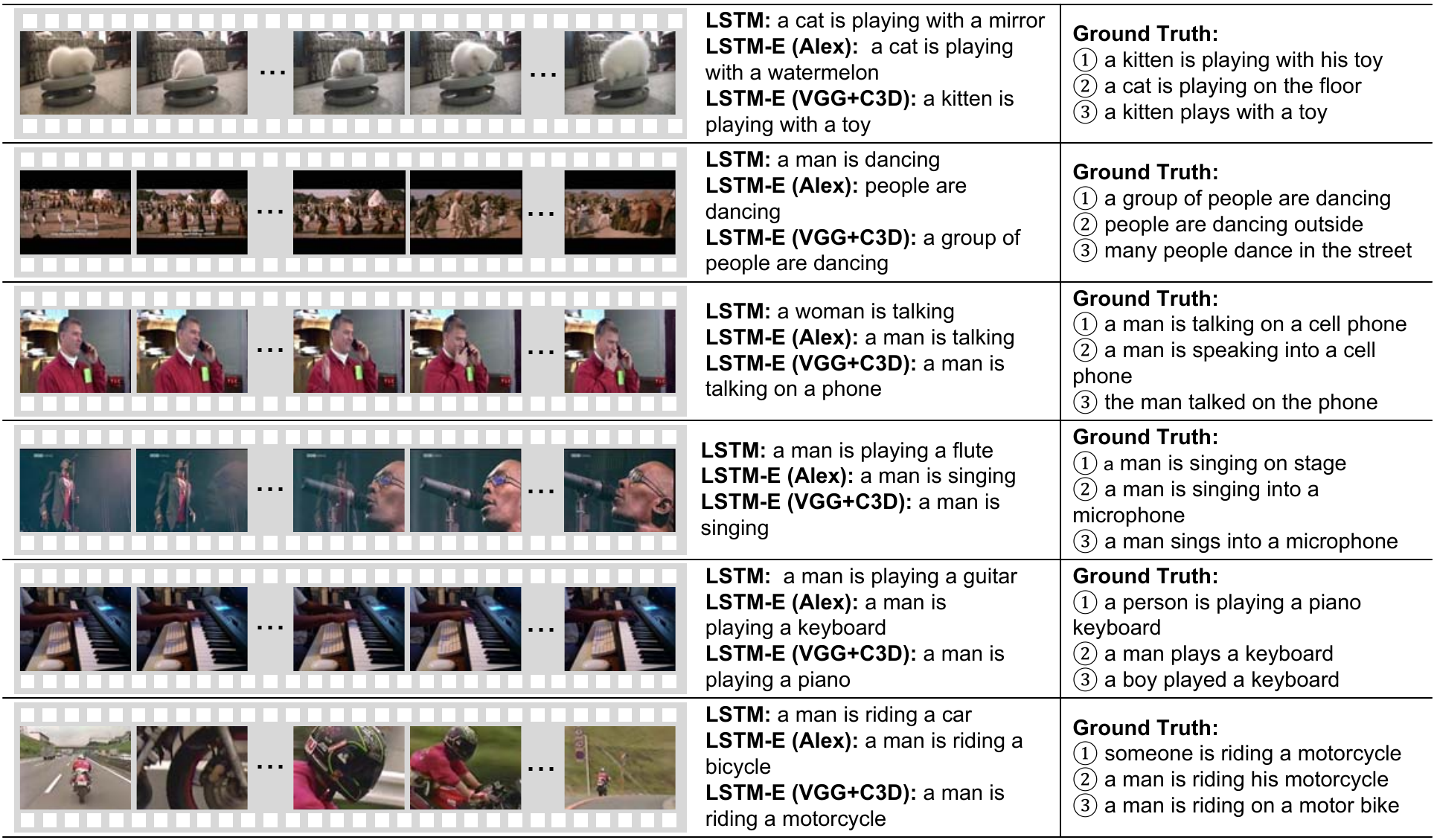}}
\caption{Examples of sentence generation results. The videos are represented by sampled frames, the output sentences generated by 1) LSTM, 2) our proposed LSTM-E (Alex) and LSTM-E (VGG+C3D), and 3) Ground Truth: Randomly selected three ground truth sentences.}
\label{fig:fig4}
\end{figure*}

Figure \ref{fig:fig4} shows a few sentence examples generated by different methods and human-annotated ground truth. From these exemplar results, it is easy to see that all of these automatic methods can generate somewhat relevant sentences. When looking into each word, both LSTM-E (Alex) and LSTM-E (VGG+C3D) predict more relevant Subject, Verb and Object (SVO) terms. For example, compared to subject term ``a man", `People" or ``a group of men" is more precise to describe the video content in the second video. Similarly, verb term ``singing" presents the fourth video more exactly. The predicted object terms ``keyboard" and ``motorcycle" are more relevant than ``guitar" and ``car" in fifth and sixth videos, respectively. Moreover, LSTM-E (VGG+C3D) can offer more coherent sentences. For instance, the generated sentence ``a man is talking on a phone" of the third video encapsulates the video content more clearly.

\newcommand{\tabincell}[2]{\begin{tabular}{@{}#1@{}}#2\end{tabular}}
\begin{table}
\centering
\caption{The effect of hidden layer size in our LSTM-E (VGG+C3D) framework measured by BLEU@4 and METEOR.}
\label{table:layer}
\begin{tabular}{c|c|c|c}\hline
\textbf{\tabincell{c}{Hidden\\layer size}} &  \textbf{BLEU@4} &  \textbf{METEOR} &  \textbf{\tabincell{c}{Parameter\\ number}} \\ \hline
128 &38.4  &29.0  &3.6M \\
256 &40.6 &29.6 &7.5M \\
512 &45.3 &31.0 &16.0M \\ \hline
\end{tabular}
\end{table}

\subsection{Experimental Analysis}\label{ssec:ExpA}
We will further provide the analysis on the effect of the tradeoff parameter between two losses and the size of hidden layer in LSTM learning.

\subsubsection{The Tradeoff Parameter $\lambda$}\label{sssec:TPS}
To clarify the effect of the tradeoff parameter $\lambda$ in Eq.(\ref{Eq:Eq12}), we illustrate the performance curves with a different tradeoff parameter in Figure \ref{fig:fig5}. To make all performance curves fall into a comparable scale, all BLEU@$N$ and METEOR values are specially normalized as follows
\begin{equation}\label{Eq:Eq14}
m_\lambda' = \frac{{{m_\lambda } - \mathop {\min }\limits_\lambda  \left\{ {{m_\lambda }} \right\}}}{{\mathop {\min }\limits_\lambda  \left\{ {{m_\lambda }} \right\}}}
\end{equation}
where $m_\lambda$ and $m_\lambda'$ denotes original and normalized performance values (BLEU@$N$ or METEOR) with a set of $\lambda$, respectively.

From the figures, we can see that all performance curves are like the ``$\wedge$" shapes when $\lambda$ varies in a range from 0.1 to 0.9. The best performance is achieved when $\lambda$ is about 0.7. This proves that it is reasonable to jointly learn the visual-semantic embedding space in the deep recurrent neural networks.

\subsubsection{The Size of hidden layer of LSTM}\label{sssec:LAY}
In order to show the relationship between the performance and hidden layer size of LSTM, we compare the results of the hidden layer size in the range of 128, 256, and 512. The results shown in Table \ref{table:layer} indicate increasing the hidden layer size can lead to the improvement of the performance with respect to both BLEU@4 and METEOR. Therefore, in our experiments, the hidden layer size is empirically set to 512, which achieves the best performance.

\section{Discussion and Conclusion}\label{sec:CON}
In this paper, we have proposed a solution to the video description problem by introducing a novel LSTM-E model structure. In particular, a visual-semantic embedding space is additionally incorporated into LSTM learning. In this way, a global relationship between the video content and sentence semantics is simultaneously measured in addition to the local contextual relationship between the word at each step and the previous ones in LSTM learning. On a popular video description dataset, the results of our experiments demonstrate the success of our approach, outperforming the current state-of-the-art model with a significantly large margin on both SVO prediction and sentence generation.

Our future works are as follows. First, as a video itself is a temporal sequence, the way of better representing the videos by using RNN will be further explored. Moreover, the video description generation might be significantly boosted if we could have sufficient labeled video-sentence pairs to train a deeper RNN.

{\small
\bibliographystyle{ieee}
\bibliography{egbib}
}

\end{document}